\documentclass{article}
\usepackage{spconf,amsmath,graphicx}
\usepackage{amssymb}
\usepackage{booktabs}
\usepackage{multicol}
\usepackage{multirow}
\usepackage{makecell}
\usepackage{setspace}
\usepackage{bm}
\usepackage{caption}
\title{learning semantic-aligned feature representation for text-based person search}
\name{{Shiping Li, Min Cao\sthanks{Corresponding author
}, Min Zhang }}

\address{School of Computer Science and Technology, Soochow University, China}
\begin{document}
%
\maketitle

\begin{abstract}
Text-based person search aims to retrieve images of a certain pedestrian by a textual description. The key challenge of this task is to eliminate the inter-modality gap and achieve the feature alignment across modalities. In this paper, we propose a semantic-aligned embedding method for text-based person search, in which the feature alignment across modalities is achieved by automatically learning the semantic-aligned visual features and textual features. First, we introduce two Transformer-based backbones to encode robust feature representations of the images and texts. Second, we design a semantic-aligned feature aggregation network to adaptively select and aggregate features with the same semantics into part-aware features, which is achieved by a multi-head attention module constrained by a cross-modality part alignment loss and a diversity loss. Experimental results on the CUHK-PEDES and Flickr30K datasets show that our method achieves state-of-the-art performances.
\end{abstract}

\section{Introduction}
\label{sec:intro}
Text-based person search~\cite{li2017person} aims to search for the corresponding person images from a large-scale image database given a textual description. Its challenge lies in two aspects: feature extraction from both visual and textual modalities and cross-modal alignment. First, it is challenging to extract robust feature representations from both images and texts due to background clutter, pose/viewpoint variances in the images, and the complexity of natural language. Then, it is difficult to overcome the cross-modal gap for alignment.
\par
Various text-based person search methods have been proposed in recent years. We generally categorize them into global-matching methods and local-matching methods. Global-matching methods \cite{wang2019language,zheng2020dual,li2017person,li2017identity,zhang2018deep} extract the global representation of samples from the two modalities separately and design proper objective functions to explore a shared latent embedding space, in which the matching scores for image-text pairs can be computed directly. However, the global-matching methods cannot effectively explore the distinctive local details of samples, which are beneficial to improving the performance. 
\par To further mine discriminative and comprehensive information, local-matching methods are proposed. The existing local-matching methods \cite{wang2020vitaa,zheng2020hierarchical,wangtext,aggarwal2020text,jing2020pose,niu2020improving,chen2018improving} generally consist of two procedures: local feature extraction and cross-modal alignment. (1) For local feature extraction, the local units in images and texts are firstly obtained by pre-set rules, based on which the local features from the local units are explicitly extracted. Specifically, some methods \cite{niu2020improving,zheng2020hierarchical} obtain the local units by simply dividing the image or its feature map into stripes or patches, and dividing the text into words, and then compute the local features by directly extracting feature representations from these units. This way introduces background noise into the features, which heavily influences the following alignment phase and causes sub-optimal retrieval results.  For this reason, some works leverage additional models, such as human parsing \cite{wang2020vitaa}, pose estimation \cite{jing2020pose}, and attribute recognition \cite{aggarwal2020text}, to locate the semantic parts in image and text as local units, yet with a heavy computing burden and non end-to-end architecture. (2) For cross-modal alignment, most local-matching methods \cite{jing2020pose,niu2020improving,10.1145/3343031.3350991,chen2018improving} adopt the cross-modality attention 
mechanism to explore the alignment between visual local units and textual ones.
\begin{figure*}
\setlength{\abovecaptionskip}{0.5cm}
\setlength{\belowcaptionskip}{-0.5cm}
\centerline{\includegraphics[width=1.0\linewidth]{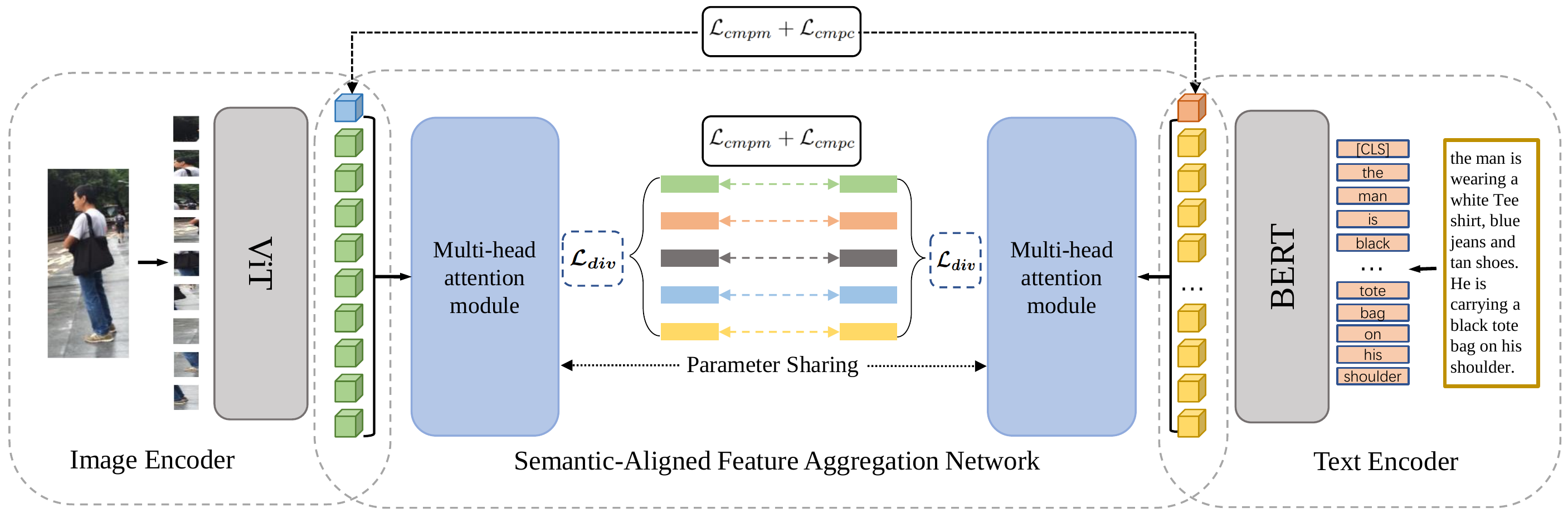}}
\vspace{-1em}\caption{The overall architecture of the proposed method. }

\label{fig:5}       
\end{figure*}
\par
To sum up, global-matching methods roughly learn a
global representation by jointly embedding the images and
texts into a shared space. Local-matching methods align
the local units through a cross-modality attention mechanism and the local units are obtained by simple dividing operations or
extra models. Compared with global-matching methods, local-matching methods can significantly boost performance owing to the information exploration at a fine-grained level and the information interaction between modalities. However, the information interaction in local-matching methods inevitably brings about efficiency damage at inference and is hardly practical in real-life applications. Thus, it is necessary to develop a simple but effective
method for text-based person search, which this paper focuses on.
\par
Motivated by the above observation, we propose a novel semantic-aligned embedding method for text-based
person search, in which the semantic-aligned part-aware features across modalities can be automatically obtained without extra model intervention and cross-modality attention mechanism. Our main contributions are three folds: 
(1) We propose a semantic-aligned feature aggregation network, which adaptively aggregates unit features with the same semantic into diverse part-aware features.
(2) To the best of our knowledge, we are the first to introduce transformer-based backbones ViT \cite{dosovitskiy2021an}, and BERT \cite{devlin2018bert} into both visual modality and textual modality to extract robust feature representations in the text-based person search.
(3) The experimental results on CUHK-PEDES \cite{li2017person}, and Flickr30K \cite{young2014image} achieve state-of-the-art performances, which verifies the effectiveness and generalization of the proposed method.

\section{METHOD}
Fig.\ref{fig:5} shows an overview of the proposed method that includes the modality-specific feature encoders (\emph{i.e.}, an image encoder and a text encoder) and a semantic-aligned feature aggregation network.

\subsection{Modality-specific Feature Encoder}
  \textbf{Image Encoder.} We adopt a ViT \cite{dosovitskiy2021an} pretrained on ImageNet \cite{deng2009imagenet} as the image feature encoder. Given an image, we first split it into a sequence of $N$ fixed-sized patches grid-likely, and then map the patch sequence to $d$ dimensional embeddings by a trainable linear projection. An extra learnable [IMG] embedding token is prepended to the sequence of patch embeddings to learn global representation. Then, we feed the patch embeddings into Transformer encoder. The output is denoted as $
 \bm{\mathrm{E}}=\left\{\bm{{\mathrm{{e}}}}_{g}, \bm{\mathrm{e}}_1, \cdots, \bm{\mathrm{e}}_{N}\right\}\in \mathbb{R}^{(N+1)\times d}$, where $\bm{\mathrm{{e}}}_g$ is the global feature of the input image, $\left\{\bm{\mathrm{e}_1}, \cdots, \bm{\mathrm{e}}_N\right\}$ are patch features.\\
  \textbf{Text Encoder.}  Given a textual description with $M$ words, we feed the sequence of words to a pretrained BERT \cite{devlin2018bert} to extract the textual features $\bm{\mathrm{T}}=\left\{\bm{\mathrm{{t}}}_{g},\bm{\mathrm{t}}_1,\cdots,\bm{\mathrm{t}}_{M}\right\}\in \mathbb{R}^{(M+1)\times d}$, where $\bm{\mathrm{{t}}}_g$ is the global feature of the input text from an extra [CLS] token, $\left\{\bm{\mathrm{t}}_1, \cdots,\bm{\mathrm{t}}_{M}\right\}$ are word features.
\subsection{Semantic-Aligned Feature Aggregation Network}
The similarity of the image-text pair can be computed by aligning the patches in the image and the words in the text, \emph{i.e.}, direct comparison between $\bm{\mathrm{E}}$ and $\bm{\mathrm{T}}$. However, such similarity is unreliable for text-based
person search due to background noise in patches and words. For this reason, we compute the similarity by aligning the image regions and the corresponding phrases in the text at the semantic level. At that point, we propose a semantic-aligned feature aggregation network, in which the region-phrase alignment is achieved by using a multi-head attention \cite{vaswani2017attention} module constrained by a cross-modality part alignment loss and a diversity loss.  

\par 
\textbf{Multi-head attention module.} 
For the visual modality, the multi-head attention module takes $\bm{\mathrm{E}}$ as input and output ${K}$ embeddings, each of which is a weighted sum of patch features. 
Concretely, inputting the visual features $\bm{\mathrm{E}}$, we firstly calculate three vectors in $i$-th head ($i=1,\cdots,K$): query $\bm{\mathrm{Q}}_i$, key $\bm{\mathrm{K}}_i$, and value $\bm{\mathrm{V}}_i$ through the linear projections,
\begin{equation}
\bm{\mathrm{Q}}_i = \bm{\mathrm{E}}\bm{\mathrm{W}}_i^Q, \bm{\mathrm{K}}_i = \bm{\mathrm{E}}\bm{\mathrm{W}}_i^K, \mathrm{V}_i = \bm{\mathrm{E}}\bm{\mathrm{W}}_i^V, 
\end{equation}
where the trainable parameter matrices $\bm{\mathrm{W}}_i^Q, \bm{\mathrm{W}}_i^K,\bm{\mathrm{W}}_i^V\in \mathbb{R}^{d\times d}$. Then we compute the $i$-th attention weight matrix $\bm{\mathrm{A}}_i \in \mathbb{R}^{(N+1)\times (N+1)}$ of the input image as
\begin{equation}
    \bm{\mathrm{A}}_i=\mathrm{softmax}(\frac{\bm{\mathrm{Q}}_i\bm{\mathrm{K}}_i^T}{\sqrt{d}}),
\end{equation}
Based on this, we obtain $K$ embeddings through
\begin{equation}
    \bm{\mathrm{E}}_i=\bm{\mathrm{A}}_i\bm{\mathrm{V}}_i.
\end{equation}
In $i$-th head, $\bm{\tilde{\mathrm{e}}}_i=\bm{{\mathrm{E}}}_i(0,:)\in \mathbb{R}^d$ represents the global representation encoded by this head, which is a weighted sum of patch features. Therefore, we obtain the visual embedding set $\bm{\tilde{\mathrm{E}}}=\left\{\bm{\tilde{\mathrm{e}}}_1,\cdots,\bm{\tilde{\mathrm{e}}}_K\right\}\in \mathbb{R}^{K \times d}$ containing $K$ feature representations of the input image encoded by different heads.\par

In the textual modality, we input the \bm{\mathrm{T}} to the multi-head attention module that shares the same parameters with that in the visual modality, and output $K$ textual embeddings
 $\bm{\tilde{\mathrm{T}}}=\left\{\bm{\tilde{\mathrm{t}}}_1,\cdots,\bm{\tilde{\mathrm{t}}}_K\right\}\in \mathbb{R}^{K \times d}$.\par
\textbf{Cross-modality Part Alignment.} 
To achieve region-phrase semantic alignment, we constrain a one-to-one alignment relationship between visual embeddings $\bm{\tilde{\mathrm{E}}}$ and textual embeddings $\tilde{\bm{\mathrm{T}}}$ by introducing a cross-modal part alignment loss, including a Cross-Modal Projection Matching (CMPM) loss and a Cross-Modal Projection Classification (CMPC) loss~\cite{zhang2018deep}.\par 

For a batch containing $n$ image-text pairs, through the multi-head attention module, we obtain $K$ visual embedding matrices denoted by $\left\{\bm{\tilde{\mathrm{\mathcal{E}}}}_1,\cdots,\bm{\tilde{\mathrm{\mathcal{E}}}}_K\right\}$, where $\bm{\tilde{\mathrm{\mathcal{E}}}}_k \in \mathbb{R}^{n\times d}$, and $K$ textual embedding matrices denoted by $\left\{\bm{\tilde{\mathrm{\mathcal{T}}}}_1,\cdots,\bm{\tilde{\mathrm{\mathcal{T}}}}_K\right\}$.
The cross-modality part alignment loss is defined as
\begin{equation}
\mathcal{L}_{part}=\sum_k^K(\mathcal{L}_{cmpm}(\bm{\tilde{\mathrm{\mathcal{E}}}}_k,\bm{\tilde{\mathrm{\mathcal{T}}}}_k)+\mathcal{L}_{cmpc}(\bm{\tilde{\mathrm{\mathcal{E}}}}_k,\bm{\tilde{{\mathrm{\mathcal{T}}}}}_k)),
\end{equation}
where the $\mathcal{L}_{cmpm}$ utilizes the KL divergence to minimize the similarities between texts and images with the different identities while maximizing the similarities between texts and images with the same identity, and the $\mathcal{L}_{cmpc}$ encourages the feature representations of samples with the same identity to be similar and discriminate from feature representations of other samples with the different identity. 
As a result, the visual feature $\bm{\tilde{\mathrm{e}}}_k$ and the textual feature $\bm{\tilde{\mathrm{t}}}_k$ with the same identity in $k$-th head contain the same semantic information.
\par\textbf{Diversity Regularization.} 
Considering that different head attention blocks could capture the redundant and overlapped semantic information to each other in the multi-head attention module, we take a further step to introduce a diversity loss $\mathcal{L}_{div}$ that penalizes the redundancy,
\begin{equation}
\mathcal{L}_{div}=\frac{1}{K(K-1)}\sum_{i=1}^K\sum_{j=1,i\neq j}^K(\frac{\bm{\tilde{\mathrm{e}}}_{i}\cdot \bm{\tilde{\mathrm{e}}}_{j}} {\Vert \bm{\tilde{\mathrm{e}}}_{i} \Vert_2\Vert \bm{\tilde{\mathrm{e}}}_{j} \Vert_2}+\frac{\bm{\tilde{\mathrm{t}}}_{i}\cdot \bm{\tilde{\mathrm{t}}}_{j}} {\Vert \bm{\tilde{\mathrm{t}}}_{i} \Vert_2\Vert \bm{\tilde{\mathrm{t}}}_{j} \Vert_2})
\end{equation}
\par
Therefore, the embeddings in $\bm{\tilde{\mathrm{E}}}$ and $\bm{\tilde{\mathrm{T}}}$ can represent different part of the input image and the input text, respectively. Through the above cross-modality part alignment loss and diversity loss, we obtain diverse semantic-aligned part-aware features from the image and text, \emph{i.e.}, $\bm{\tilde{\mathrm{E}}}$ and $\bm{\tilde{\mathrm{T}}}$.\par
\subsection{Training and inference}
The entire network is trained in an end-to-end manner. As a supplementary to the cross-modal part alignment, we add a cross-modal global alignment. We denote the global features of images and texts in a batch as $\bm{\mathrm{\mathcal{E}}}_g \in \mathbb{R}^{n\times d}$ and $\bm{\mathrm{\mathcal{T}}}_g \in \mathbb{R}^{n\times d}$, respectively. The overall loss is computed by
\begin{equation}
\mathcal{L}=\mathcal{L}_{global}+\mathcal{L}_{part}+\lambda\mathcal{L}_{div},
\label{eq6}
\end{equation}
where $
\mathcal{L}_{global}=\mathcal{L}_{cmpm}(\bm{\mathrm{\mathcal{E}}}_g,\bm{\mathrm{\mathcal{T}}}_g)+\mathcal{L}_{cmpc}(\bm{\mathrm{\mathcal{E}}}_g,\bm{\mathrm{\mathcal{T}}}_g)$ and 
$\lambda$ is a parameter to control the importance of the diversity loss.
\par
During the inference, the similarity score between one image-text pair is measured as the sum of the cosine distance of the global features and part-aware features between them,
\begin{equation}
Sim(I,T)=cosine(\bm{\mathrm{e}}_g,\bm{\mathrm{t}}_g)+\sum_k^Kcosine(\bm{\tilde{\mathrm{e}}}_k,\bm{\tilde{\mathrm{t}}}_k).
\end{equation}
\section{EXPERIMENTS}
\subsection{Datasets and implementation details}
We conduct experiments on the only text-based person search benchmark CUHK-PEDES \cite{li2017person}, which contains 40,206 images of 13,003 identities, with 2 captions per image, 11,003 for training, 1,000 for evaluation, and 1,000 for test. To verify the generalization of the proposed method, we also conduct experiments on a cross-modal retrieval dataset Flickr30K \cite{young2014image},  in which there are 31,784 images with five captions for each in total. We take the same split for training, validation and testing set as \cite{karpathy2015deep}. Flickr30K contains a variety of objects, rather than only pedestrians as CUHK-PEDES. \par
In the experiments, we resize the image to $224\times224$ and pad the text to length 100. There are 50 epochs in training, and the batch size is set to 64. The dimension of visual and textual features is set to $d=768$. The parameter $K$ for the multi-head attention module is set to $10$ and the parameter $\lambda$ in Eq. \ref{eq6} is set to $0.2$. Following the standard setting \cite{wang2020vitaa}, we adopt Rank-1/5/10 as the evaluation criteria.
\subsection{Comparisons with State-of-the-art Methods}
\textbf{Results on CUHK-PEDES Dataset.}
The comparison results in Table \ref{tab1} show that the proposed method performs the best results at Rank-1/5/10 accuracies with all comparisons. Particularly, the proposed method outperforms the second-best method SSAN \cite{ding2021semantically} by a large margin, such as a 2.76\% gain at Rank-1.
It is worth noting that the compared methods GLA \cite{chen2018improving}, PMA\cite{jing2020pose} and MIA \cite{niu2020improving} construct the elaborate cross-modality attention mechanism for achieving cross-modal alignment with low efficiency. By contrast, the proposed method reaches overbearing advantages on performance by performing an automatic cross-modal alignment with a simple and lightweight architecture.
\begin{table}
\begin{center}
\setlength{\abovecaptionskip}{0.5cm}
\caption{Performance comparison with state-of-the-arts on CUHK-PEDES dataset.}
\label{tab1}
\vspace{-1em}
\begin{tabular}{c|c|c|c}
\Xhline{1.2pt}
Method & Rank-1 & Rank-5 & Rank-10 \\
\hline
GNA-RNN \cite{li2017person} & 19.05 & ~~- & 53.64 \\
GLA \cite{chen2018improving}& 43.58&66.93 &76.26\\
Dual-path \cite{zheng2020dual} & 44.40 & 66.26 & 75.07 \\
CMPM+CMPC \cite{zhang2018deep} & 49.27 & ~~- & 79.27 \\
MCCL \cite{wang2019language} &50.58 &-&79.06\\
MIA \cite{niu2020improving} & 53.10 & 75.00 & 82.90 \\
A-GANet \cite{10.1145/3343031.3350991} & 53.14 &74.03 & 81.95\\
PMA \cite{jing2020pose} & 53.81 & 73.54 & 81.23 \\
TIMAM \cite{zha2020adversarial}& 54.51 & 77.56 & 84.78 \\
ViTAA \cite{wang2020vitaa} & 55.97 & 75.84 &83.52 \\
MGEL \cite{wangtext} & 60.27 & 80.01 & 86.74 \\
SSAN \cite{ding2021semantically} & 61.37 & 80.15 & 86.73 \\
\hline
Ours &\textbf{64.13} & \textbf{82.62} & \textbf{88.40} \\
\Xhline{1.2pt}
\end{tabular}
\end{center}
\vspace{-1em}
\end{table}

\par
\textbf{Results on Flickr30K Dataset: } We compare the proposed method against several state-of-the-art methods in Table \ref{tab2}. It can be shown that the proposed method surpasses all other methods at all Rank accuracies, showing the effectiveness and generalization of the proposed method.
\begin{table}

\setlength{\abovecaptionskip}{0.5cm}
\begin{center}
\caption{Performance comparison with state-of-the-arts on Flickr30K dataset.}
\label{tab2}
\vspace{-1em}
\begin{tabular}{c|c|c|c}
\Xhline{1.2pt}
\multirow{2}*{Method} & \multicolumn{3}{c}{Text-to-Image} \\
\cline{2-4}
& Rank-1 & Rank-5 & Rank-10 \\

\hline
Dual-path \cite{zheng2020dual} & 39.10 & 69.20 & 80.90 \\
CMPM+CMPC \cite{zhang2018deep} & 37.30 & 65.70 & 75.50 \\
A-GANet \cite{10.1145/3343031.3350991} &39.52 & 69.91 & 80.91 \\
TIMAM \cite{zha2020adversarial}& 42.60 & 71.60 & 81.90 \\
\hline
Ours &\textbf{ 50.74} & \textbf{77.92} & \textbf{85.46} \\
\Xhline{1.2pt}
\end{tabular}
\end{center}
\vspace{-2em}
\end{table}

\par
\subsection{Ablation Studies}
\textbf{Analysis on modules of the proposed method:} 
There are two modules in the proposed method: the modality-specific feature encoder (MSFE) and the semantic-aligned feature aggregation network (SAFA). The global matching from the first module and the local matching from the second one is incorporated into the matching computation for text-based person search. We analyze the impacts of the two modules on performance. The results are shown in Table \ref{tab3}.  Compared to Ours$_g$, Ours achieves performance enhancement, (\emph{i.e.}, 4.34\%, 3.02\% and 2.39\% gains at Rank-1/5/10), which demonstrates the effectiveness of the proposed local-matching from the SAFA module. In addition, Ours obtains the minor increases at Rank-1/5 compared with ours$_p$, indicating that the global-matching from the MSFE module is an effective auxiliary for the cross-modal matching.
\begin{table}
\setlength{\abovecaptionskip}{0.5cm}
\setlength{\belowcaptionskip}{-0.5cm}
\begin{center}
\caption{Ablation study of the proposed method on CUHK-PEDES dataset.}\label{tab3}
\vspace{-1em}
\begin{tabular}{c|c|c|c|c|c}
\Xhline{1.2pt}
Method & global & part & Rank-1& Rank-5& Rank-10 \\
\hline
Ours$_g$ & \checkmark & & 59.80&79.60&86.01 \\
Ours$_p$ &  & \checkmark &63.90&82.50&\textbf{88.71} \\
Ours &\checkmark & \checkmark  &\textbf{64.13}&\textbf{82.62}&{88.40}\\
\Xhline{1.2pt}
\end{tabular}
\end{center}
\vspace{-1.5em}
\end{table}
\begin{figure}
\setlength{\abovecaptionskip}{0.5cm}
\setlength{\belowcaptionskip}{-0.5cm}
\centering
\includegraphics[width=8.5cm]{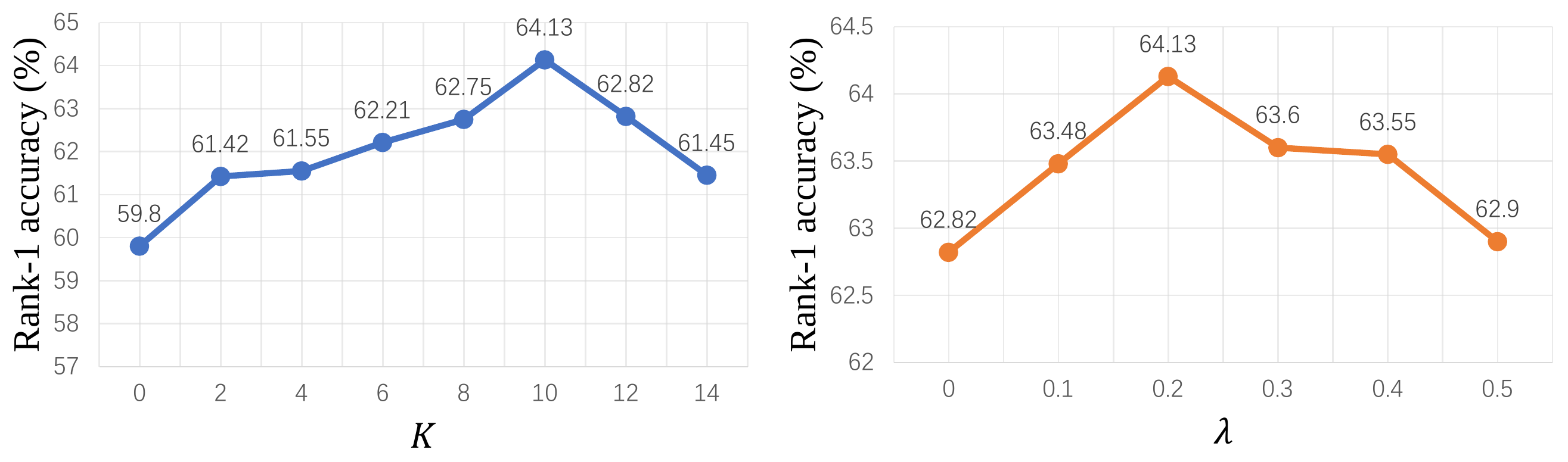}
\vspace{-1.5em}
\caption{Influence of the parameters in the proposed method on CUHK-PEDES.}

\label{fig:1} 
\end{figure}

\par
\textbf{Parameters analysis:} There are two parameters $K$ and $\lambda$ in the proposed method. Fig.\ref{fig:1} shows the impact of two parameters at Rank-1. (1) For parameter $K$, the best result is achieved with $K=10$. A smaller value to $K$ represents insufficient exploration on the fine-grained region-phrase alignment and results in performance degradation. Besides, a larger value to $K$ may introduce noise information for alignment and lead to the decrease of performance. (2) For parameter $\lambda$, the proposed method with $\lambda=0.2$ performs the best result.
We can study that when setting $\lambda>0$, the proposed method always achieves performance enhancement, showing the effectiveness of the diversity loss for performance.
\subsection{Visualization Analysis}
\begin{figure}

\setlength{\abovecaptionskip}{0.5cm}
\setlength{\belowcaptionskip}{-0.5cm}
\vspace{1em}
\begin{spacing}{0.6}
{\scriptsize \textbf{Caption:} The man wears glasses, has short, graying hair, is holding a briefcase in his right hand, and is wearing a dark jacket and dark pants.
}
\centering
\label{vv}
\includegraphics[width=1\linewidth, height=0.62\linewidth]{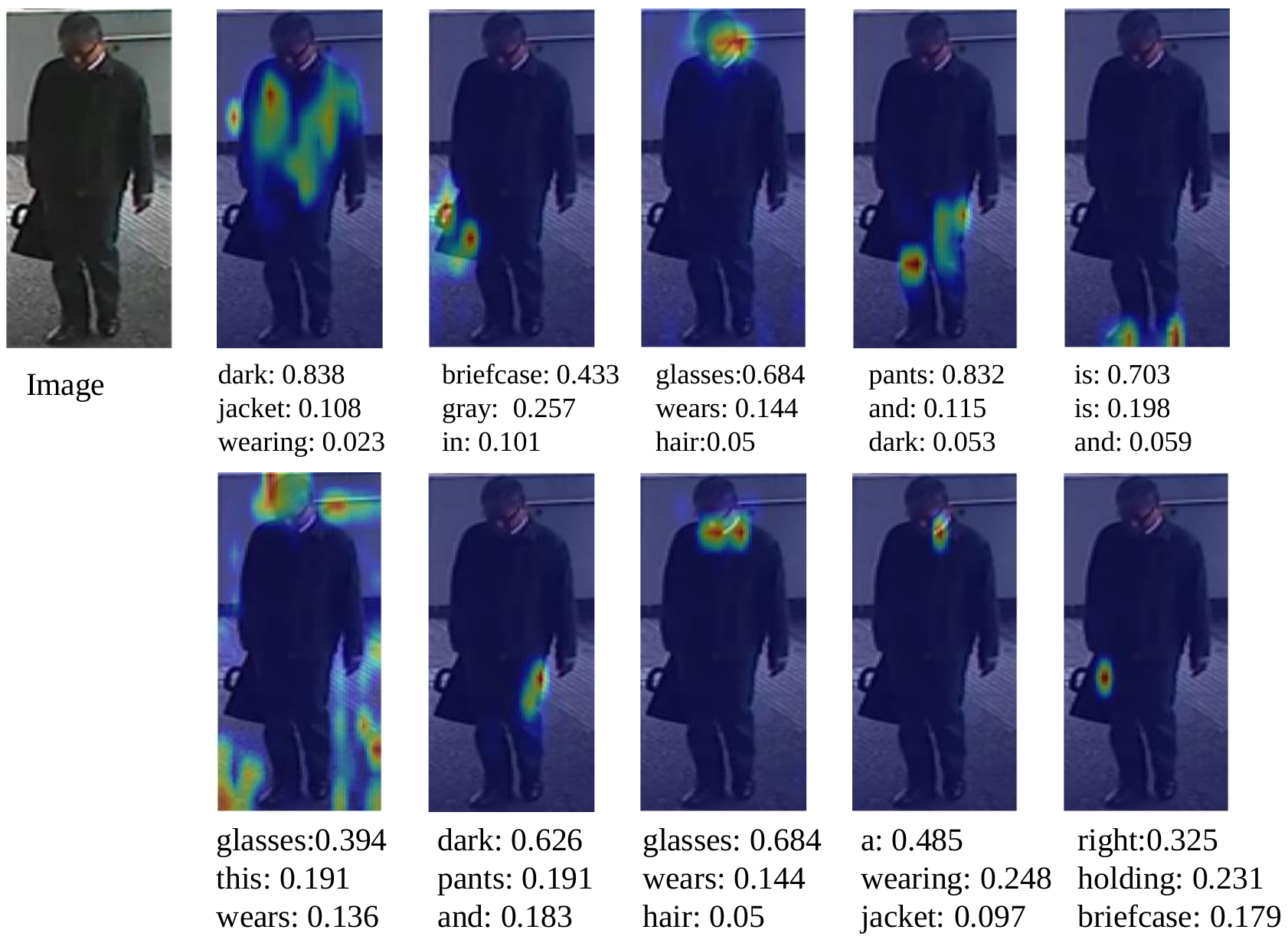}
\vspace{-1em}\caption{Visualization of the semantic alignment across modalities by the proposed method.}
\end{spacing}
\vspace{0.5em}
\end{figure}
We visualize the semantic alignment across modalities from the multi-head attention module in the proposed method. Specifically, we show the attention map of each head in the image and the words with the first three attention values among all the words in the text in Fig.3. We can observe that each head attends to a different part of the image and the text, and the part's semantic information from them is interrelated with each other. It shows that the proposed method achieves effective semantic alignment across modalities.
\section{CONCLUSIONS}
\label{sec:conclusions}
\par In this paper, we propose a simple but effective method for text-based person search. In contrast to the existing local-matching methods, the proposed method is an end-to-end trainable architecture without additional models and complex cross-modal information interaction strategies. We design a semantic-aligned feature aggregation network for learning the part-aware features that are semantic-aligned across modalities adaptively. The experimental results on CUHK-PEDES and Flickr30K verify the superiority of the proposed method.

\bibliographystyle{IEEEbib}
\bibliography{strings,refs}

\end{document}